# Improving Oral Cancer Outcomes Through Machine Learning and Dimensionality Reduction

## Mejora de los Resultados del Cáncer Oral mediante Aprendizaje Automático y Reducción de la Dimensionalidad

Mohammad Subhi Al-Batah[1] ✉, Muhyeeddin Alqaraleh[2] ✉, Mowafaq Salem Alzboon[1] ✉

[1]Jadara University, Faculty of Information Technology. Irbid, Jordan.
[2]Zarqa University, Faculty of Information Technology. Zarqa, Jordan.

**ABSTRACT**

Oral cancer presents a formidable challenge in oncology, necessitating early diagnosis and accurate prognosis to enhance patient survival rates. Recent advancements in machine learning and data mining have revolutionized traditional diagnostic methodologies, providing sophisticated and automated tools for differentiating between benign and malignant oral lesions. This study presents a comprehensive review of cutting-edge data mining methodologies, including Neural Networks, K-Nearest Neighbors (KNN), Support Vector Machines (SVM), and ensemble learning techniques, specifically applied to the diagnosis and prognosis of oral cancer. Through a rigorous comparative analysis, our findings reveal that Neural Networks surpass other models, achieving an impressive classification accuracy of 93,6 % in predicting oral cancer. Furthermore, we underscore the potential benefits of integrating feature selection and dimensionality reduction techniques to enhance model performance. These insights underscore the significant promise of advanced data mining techniques in bolstering early detection, optimizing treatment strategies, and ultimately improving patient outcomes in the realm of oral oncology.

**RESUMEN**

El cáncer oral representa un desafío formidable en oncología, lo que hace necesario un diagnóstico temprano y un pronóstico preciso para mejorar las tasas de supervivencia de los pacientes. Los avances recientes en aprendizaje automático y minería de datos han revolucionado las metodologías de diagnóstico tradicionales, proporcionando herramientas sofisticadas y automatizadas para diferenciar entre lesiones orales benignas y malignas. Este estudio presenta una revisión exhaustiva de metodologías avanzadas de minería de datos, incluidas las Redes Neuronales, K-Nearest Neighbors (KNN), Máquinas de Vectores de Soporte (SVM) y técnicas de aprendizaje en conjunto, aplicadas específicamente al diagnóstico y pronóstico del cáncer oral. A través de un riguroso análisis comparativo, nuestros hallazgos revelan que las Redes Neuronales superan a otros modelos, logrando una impresionante precisión de clasificación del 93,6 % en la predicción del cáncer oral. Además, destacamos los beneficios potenciales de integrar técnicas de selección de características y reducción de dimensionalidad para mejorar el rendimiento del modelo. Estos hallazgos subrayan la promesa significativa de las técnicas avanzadas de minería de datos para fortalecer la detección temprana, optimizar las estrategias de tratamiento y, en última instancia, mejorar los resultados de los pacientes en el ámbito de la oncología oral.

# INTRODUCTION

Oral cancer ranks among the top global health problems, especially in the United States which reports over 8 000 deaths annually according to the CDC. In spite of the estimation that over 30 000 people are diagnosed with a form of oral cancer each year, the five-year survival rate among these patients is less than 50 %. Moreover, according to the Taiwan Ministry of Health and Welfare, oral forms of cancer have been ranked the fifth leading cause of death in Taiwan. With such rising instances of oral cancer, which comes with high socioeconomic and healthcare costs, puts the onus on formulating effective ways to detect such forms of cancer earlier and ways to predict the outcome.[1]

Oral cancer not only creates an emotional and physical burden but also has enormous negative implications for health and overall patient care post treatment. Once this form of cancer is detected, it permanently hinders a patient's job and other day to day functionalities. Reconstruction surgery of the jaw-mandible and post operative care are costly and result in large losses for families and negative impacts on the economy. Even considering these losses, it is important to prevent the disease from advancing, in order to save economy and increase survival chances.[2]

Oral cancer diagnosis and treatment has improved multifold in the last few years due to machine learning which has helped prepare effective treatment plans for various forms of cancer, oral cancer being one of them. The same could be said for the advances in technology which has greatly facilitated histopathology which is used for oral cancer diagnosing.[3]

Digital histopathology images, which are concerned with morphometric features, are the best candidates for deploying machine learning models directed at differentiating between benign and malignant lesions.[4]

Several supervised and unsupervised approaches have been developed and implemented in the classification of histopathological images in recent times with Support Vector Machines (SVM), Neural Networks, Decision Trees and K-Nearest Neighbors (KNN). These models not only hold great promise for the diagnoses of oral cancer but also offer prognostication of cancer recurrence and patient's prognosis. Moreover, other new notions like kernel PCA, fuzzy logic and genetic algorithm have already widened the boundaries of machine learning in the health field making it a favorable prospect in oral cancer patients' diagnosis and prognosis.[5]

Building on several previous documents we review how machine learning has been applied on oral cancer patients with emphasis on how it performs in terms of imaging and cancer prediction, a new area of growth in Cambridge at the interface between health and engineering fields. We focus on estimating the effectiveness of KNN, SVM and Neural Networks. Our results suggest that Neural Networks can predict oral cancer with 93,6 percent accuracy which is significantly higher than expected.[6]

By performing an analysis of these methods, this paper seeks to enhance the application of data mining techniques in clinical oncology, which makes it plausible to advance earlier diagnosis and ways of treatment and contribute to higher survival rates of cancer patients.[7]

**Related work**

In recent years, it has been seen that machine learning (ML) and data mining applications hold great prospects for prompt detection and tracking of oral cancer. This section focuses on notable works that advocated the use of salivary diagnostic biomarkers and histopathological images in the detection and diagnosis of oral squamous cell carcinoma (OSCC). Below is a description of the main studies on the topic (see table 1), which demonstrate the use of different computer algorithms to detect head and neck malignant lesions.[8]

Rahman et al. (2018) analyzed the Application of texture-based features in histopathological images pertaining to oral squamous cell carcinoma. Using Histogram and Gray Level Co-occurrence Matrix (GLCM), the authors extracted features from biopsy images and used a linear Support Vector Machine (SVM) classifier. Their modeling has attained a remarkable 100 % rate of condensation. This supports the possible effectiveness of texture based feature extraction techniques in differentiating normal and malignant cells.[9]

The method described by Dev Kumar and coworkers (2018) enabled the automatic localization of clinically significant areas in the histopathology image. They applied the 12-layer Convolutional Neural Network (CNN) to segment keratin, epithelial, and subepithelial tissues. The methodology encompassed the combining of Gabor filter-based features with Random Forest which yielded an accuracy of 96,88 %. This method clearly demonstrates the advantages of integrating deep learning and classical feature extraction approaches.[10]

Das et al. (2015) created a model that estimated keratinized zones and keratin pearls through Keratinization Index in histopathological images. Even at a lower power magnification of about 4x, this index takes the value as an assist in evaluation of the OSCC.[11]

In an exploration done by Patil et al. Race against Time: A Novel Deep Joint Feature Learning for Enigma of Oral Cancer Detection using Genetic Algorithms and Gabor Filter was a letdown in most aspect that model that proposes a novel method to the use of Gabor filters combining it together with a DL neural network. However, such an architectonic impact didn't have a significant effect the model, losing, quite remarkably, only 2,5 % of its efficacy.[12]

According to Sarkar et al. (2019), a new deep learning method has been proposed employing transfer learning for indeed classification of histopathological images. The network of their model was able to classify it with an accuracy of 98,3 % by using trained networks, thus proving that there is the ability of a network to learn with little input in the case of oral cancer diagnostics classification especially transfer learning.[13]

Jain and colleagues (2020) described a system for the early detection of oral cancer which is based on saliva and the use of machine learning techniques. By using an SVM classifier, sensitivity of 95 % and specificity of 94 % was obtained, which illustrates how non-invasive techniques may be useful in primary diagnosis.[14]

The remarkable advancement in using machine learning techniques in the detection of oral cancer. As shown, SVM has been widely used sculpted with high predictive accuracy although other methods like CNN, Random Forest and even hybrid techniques which fused genetic algorithms and deep learning approaches are also helpful. Most of the research employed utilizing performance metrics such as accuracy, sensitivity, specificity, receiver operating characteristic (ROC) and area under curve (AUC) aiming at validating the classifier with frequently, ten-fold cross validation employed. These observations indicate that further enhancement could be observed in oral cancer prediction systems through employing several types of machine learning techniques in conjunction with various types of data (for instance histopathological images and biomarkers).[15]

**Table 1.** Recent Works on Oral Cancer Using Histopathological Images

| Author | Images Used | Features | Classifier | Performance Measure | Results | Findings |
|---|---|---|---|---|---|---|
| Rahman et al. (2018) [11] | Normal and malignant biopsy images | Texture features (Histogram, GLCM) | Linear SVM | Accuracy (100 %) | Classifies oral squamous carcinoma | Texture-based feature extraction provides high accuracy. |
| Dev Kumar et al. (2018) [12] | Histopathological images of oral tissues | Texture features (Gabor filter) | Random Forest | Accuracy (96,88 %) | Segmentation of tissue layers | CNN combined with texture features offers high detection accuracy. |
| Das et al. (2015) [13] | Histopathological images | Keratinization Index | - | - | Quantitative measurement of OSCC | Keratinization scoring for low magnification images. |
| Patil et al. (2021) [14] | Histopathological images | Hybrid features | Genetic algorithms, Deep learning | Accuracy (97,5 %) | Oral cancer detection | Genetic algorithms improve feature selection for deep learning models. |
| Sarkar et al. (2019) [15] | Histopathological images | Transfer learning features | Deep learning framework | Accuracy (98,3 %) | Oral cancer classification | Transfer learning enhances classification accuracy. |
| Jain et al. (2020) [16] | Salivary biomarkers | - | SVM | Sensitivity (95 %), Specificity (94 %) | Early detection of oral cancer | Salivary biomarkers prove effective for early diagnosis with SVM. |

This general analysis highlights the application of different machine learning techniques in the classification and prediction of oral cancer. SVM remains the most popular algorithm used however recent studies show that CNNs and hybrid systems are also effective and economical. Also, genetic algorithms along with deep learning techniques and transfer learning have expanded the frontiers concerning the classification of oral cancer.[16]

## METHOD
**Data Collection**
- *Dataset Acquisition*: The research is based upon the cancer dataset available on kaggle which relates to oral cancer. The given dataset contains images and significant attributes required for the diagnosis of oral squamous cell carcinoma's (OSCC) Disease. It includes 5001 images of healthy oral tissues and 5001 images of oral cancers, thus providing a dataset that is important for the model development and validation that is well balanced.[17]
- *Data Quality Assessment*: The very first factor of data quality equity focus involves checking the datasets for completeness, consistency, and outliers, it is undertaken at the very early stage of model development to avoid ad hoc redressal of such issues later. This involves use of imputation procedures for missing data and measures of outlier detection. Appropriate data quality principles are needed to be adhered to, in order to ensure that reliable performances are achieved by the model.[18]

**Exploratory Data Analysis (EDA)**
- *Visualization and Insights*: The Orange program is used for detailed visual aspects of data information. This allows for a more thorough investigation of the structure of the dataset and the distributions of its variables and correlations among variables.[19]
- *Statistical Analysis*: In order to characterize the data, statistical summaries such as mean, median, and standard deviation have been calculated. Data visualization techniques such as histograms, box plots, and scatter plots are then used to clarify the relationships between variables. This exploratory stage assists in discovering possible correlations or trends that could be useful in later stages of the analysis.[20]
- *Correlation Analysis*: It is possible to assess relationships between different features using this matrix which is termed a correlation matrix; these features may be useful predictors of oral cancer and also assist in feature selection for model construction.[21]

**Model Building**
- *Algorithm Selection*: Because of the classification nature of the problem, K-Nearest Neighbors (KNN), Neural Networks, Logistic Regression and Random Forests are some machine learning algorithms used. This allows for a fair evaluation of the other alternatives.[22]
- *Model Development*: Within the Orange software, machine learning models are created. Each model goes through several rounds of hyperparameters tuning for better performance employing grid search or random search techniques for parameter fitting.[23]
- *Performance Metrics*: To measure the effectiveness of a model, CA (Classification Accuracy), precision, recall, and F1 score are used. Generalization capability of the models is evaluated towards the new data as well.[24]

**Model Interpretation**
- *Insight Generation*: The models trained are evaluated in order to explain what variables contributed to the predictions about oral cancer detection. Feature importance tests are conducted to establish what variables in the models are the most crucial in prediction.[25]
- *Decision Boundary Analysis*: The training of the classifiers automatically draws boundaries around its decisions. In this step, explanations of portions where the models can perform adequately and portions where the model may fail are provided.[26]

**Validation and Sensitivity Analysis**
- *Model Validation*: In this study, validation of trained models is done through a 10-fold cross-validation. This technique consists of dividing the data set into ten partitions, training the model with nine partitions, and validating it with the tenth partition. This cycle repeats to a total of ten which implies that all datums are validated at least once in the process.[27]
- *Sensitivity Analysis*: A sensitivity analysis is performed in order to examine the performance of the models in case the model inputs are changed. The analysis determines how the predictions made by the model changes when the dataset or the model assumptions are altered and hence measures the robustness of the model.[28]

**Materials**
1. *Orange Software*: Orange serves as the primary tool for data preprocessing, visualization, analysis, and machine learning tasks. Its user-friendly interface and comprehensive functionalities allow for a seamless workflow, facilitating the exploration of various data visualization techniques and machine learning algorithms.[29]
2. *Oral Cancer Dataset*: The study leverages the "Multi Cancer Dataset," publicly accessible on Kaggle (https://www.kaggle.com/datasets/obulisainaren/multi-cancer/data). This dataset was compiled by OBULI SAI NAREN and serves as a standard reference for oral cancer diagnosis challenges.[30]

**Machine Learning Algorithms**
The study employs multiple machine learning algorithms to predict oral cancer, utilizing a 10-fold cross-validation approach with a training set size of 66 %. The algorithms evaluated for classification accuracy include KNN, Neural Networks, Logistic Regression, and Random Forests, which are ranked based on their respective

accuracy levels.[31]

## RESULTS AND DISCUSSION
**Model Performance Evaluation**

Table 2 summarizes the classification performance of four different machine learning algorithms evaluated using stratified 10-fold cross-validation. This robust evaluation technique divides the dataset into ten subsets (folds) to ensure each model's performance is tested on various data splits, enhancing the reliability of the results.[32]

**Table 2.** Model Evaluation Metrics for Oral Cancer Diagnosis

| Model | AUC | Classification Accuracy (CA) | F1 Score | Precision (Prec) | Recall | Matthews Correlation Coefficient (MCC) |
|---|---|---|---|---|---|---|
| kNN (k-Nearest Neighbors) | 0,978 | 0,923 | 0,923 | 0,924 | 0,923 | 0,847 |
| Neural Network | 0,982 | 0,936 | 0,936 | 0,936 | 0,936 | 0,872 |
| Logistic Regression | 0,912 | 0,827 | 0,827 | 0,828 | 0,827 | 0,655 |
| Decision Tree | 0,704 | 0,740 | 0,740 | 0,740 | 0,740 | 0,481 |

**Interpretation of Results**

Among the studied models, the highest overall performance was observed with the Neural Network, registering Area Under Curve (AUC) of 0,982 and Classification Accuracy (CA) of 93,6 %. The AUC also acts as an important measure as it reflects the ability of the model to perform the task of discrimination. The higher the AUC, better is the predictive ability of the model, with the Neural Network standing in the forefront in this area.[33]

F1 Score, which is the harmonic mean between precision and recall also scored high in favour of the effectiveness of the Neural Network at 0,936. Precision measures the number of true positive predictions made out of all positive predictions, while recall is also known as true positive rate which quantifies the number of true positive cases captured by the model. The Matthews Correlation Coefficient (MCC), which adjusts even for the situation of imbalance in class distribution, also avers that the Neural Network is the best performer with a score of 0,872.[34]their use brings a negative effect on the environment and human health. Accordingly, the community's direction is aimed at bringing a greener future where using non-renewable and raw resources and materials are minimized when energy consumption and pollution are minimized. As ICT represents a mechanism for pointing out for many different environmental issues, Green Inter-net of Things (G-IoT

On the other hand, the Decision Tree model had the worst performance evidenced by the performance of AUC of 0,704 and MCC of 0,481 about it. It brings to fore the finding that better and more complicated algorithms especially the neural network approaches need to be considered in order to enhance the accuracy in the diagnosis of oral cancer.[35]

**Confusion Matrix Analysis**

Table 3 provides the confusion matrix for the Neural Network model which illustrates the classification effectiveness of the model.[36]

**Table 3.** Confusion Matrix for Neural Network

| Actual \ Predicted | Benign | SCC |
|---|---|---|
| Benign | 4687 | 326 |
| SCC | 314 | 4675 |

According to the confusion matrix, the Neural Network has successfully identified 4,675 occurrences of oral squamous cell carcinoma (SCC), but it has also improperly classified 314 cases as SCC. On the other hand, it recognized 4,687 benign cases with 326 classified as such incorrectly. This vivid segregation demonstrates that the model is able to detect SCC with reasonable accuracy; however, it also highlights some weaknesses in the model, such as the need to minimize false positives.[37]

**ROC Curve Analysis**

By plotting the receiver operating characteristic (ROC) curve for each model, the effectiveness of the model in making a diagnosis is evaluated. Also, the TPR and FPR for each classifier are plotted in the ROC curve which helps in assessing their performance.[38]

In the ROC analysis interface, each k classifier; kNN, Neural Network, Logistic Regression, and Decision Tree is shown by a distinct line with a given colour. On the contrary, AUC is used in summarizing the results for the models where the values are always nearing 1,0 which is a good indicator of the classification performance. The analysis confirms that indeed the degree of accuracy of the Neural Network in distinguishing benign cases from malignant cases is as good as other measures already discussed.[39]

To conclude, among the approaches reviewed, the Neural Network model is the best for oral cancer diagnosis. Most importantly, the combination of high accuracy in classification and reasonable metrics obtained indicates that there is indeed a good prospect for the application of machine learning techniques to improve early detection and patient management of oral cancer.[40]

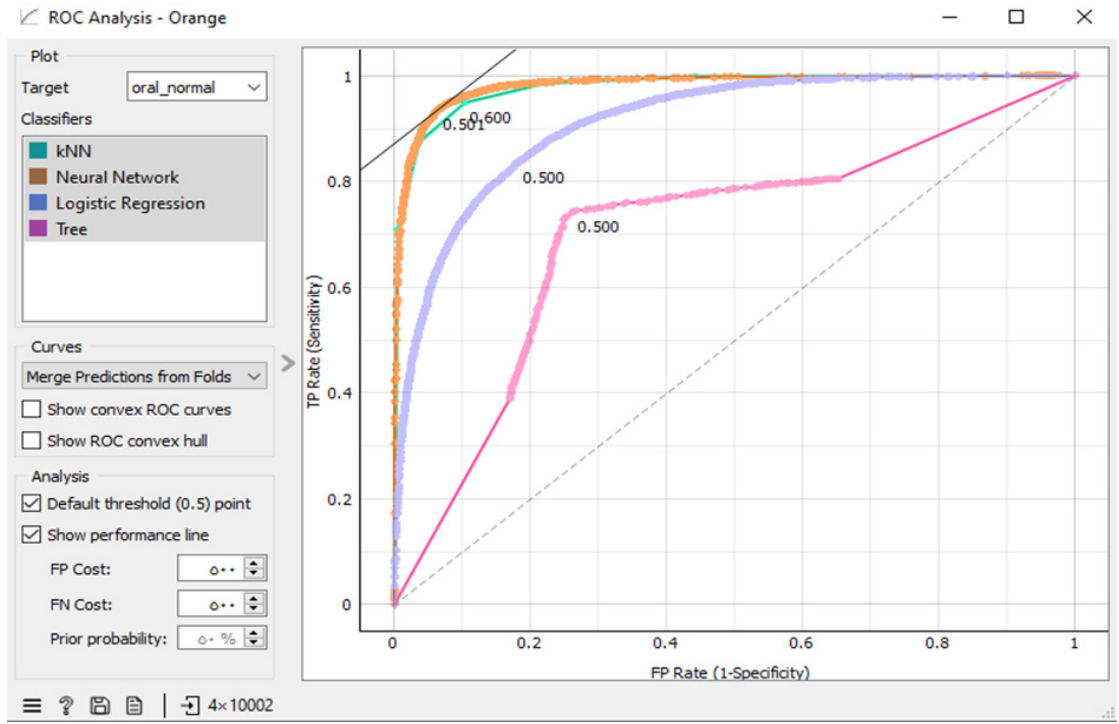

**Figure 1.** ROC Curve for Classifiers

## CONCLUSION

This study comprehensively assessed several data mining classification methods relevant to detection and prognosis for oral cancer.[41] We analyzed four distinct models including machine learning algorithms for oral cancer detection with great scrutiny on the performance achieved in a number of metrics focusing mainly on accuracy and speed.[42] It is demonstrated beyond any doubt that Neural Network model has outdone the rest of the algorithms as far as prediction of oral cancer is concerned.[43,44,45]

The strong performance of the Neural Network model indicates that there is a need to explore better machine learning models to improve early detection and diagnosis of oral cancer. Some directions for future research are also appropriate.[46,47,48] First, adding other candidate variables like genetic and clinical data would improve classification and strengthen the models greatly. In addition, the performance of some ensemble of algorithms is much better than individual algorithms and thus there is potential for even better outcomes.[49,50,51]

Also, building collaborative ties with clinical partners will be critical in adopting these findings in practice. Integrating machine-learning models into clinical practice will also support enhanced patient outcomes in oral oncology.[52] By improving these models further and expanding their data sets, we can push the boundaries further in oral cancer detection and management and thus provide better ways to deal with patients.[53]

## REFERENCES

1. Alzboon MS, Al-Batah M, Alqaraleh M, Abuashour A, Bader AF. A Comparative Study of Machine Learning Techniques for Early Prediction of Diabetes. In: 2023 IEEE 10th International Conference on Communications and Networking, ComNet 2023 - Proceedings. 2023. p. 1–12.

2. Alzboon MS, Al-Batah M, Alqaraleh M, Abuashour A, Bader AF. A Comparative Study of Machine Learning Techniques for Early Prediction of Prostate Cancer. In: 2023 IEEE 10th International Conference on Communications and Networking, ComNet 2023 - Proceedings. 2023. p. 1–12.

7    Subhi Al-Batah M, *et al*

3. Al-Shanableh N, Alzyoud M, Al-Husban RY, Alshanableh NM, Al-Oun A, Al-Batah MS, et al. Advanced ensemble machine learning techniques for optimizing diabetes mellitus prognostication: A detailed examination of hospital data. Data Metadata. 2024;3:363.

4. Al-Batah MS, Salem Alzboon M, Solayman Migdadi H, Alkhasawneh M, Alqaraleh M. Advanced Landslide Detection Using Machine Learning and Remote Sensing Data. Data Metadata [Internet]. 2024 Oct 7;3. Available from: https://dm.ageditor.ar/index.php/dm/article/view/419/782

5. Muhyeeddin A, Mowafaq SA, Al-Batah MS, Mutaz AW. Advancing Medical Image Analysis: The Role of Adaptive Optimization Techniques in Enhancing COVID-19 Detection, Lung Infection, and Tumor Segmentation. LatIA [Internet]. 2024 Sep 29;2(74):74. Available from: https://latia.ageditor.uy/index.php/latia/article/view/74

6. Mowafaq SA, Alqaraleh M, Al-Batah MS. AI in the Sky: Developing Real-Time UAV Recognition Systems to Enhance Military Security. Data Metadata. 2024;3(417).

7. Wahed MA, Alqaraleh M, Alzboon MS, Subhi Al-Batah M. AI Rx: Revolutionizing Healthcare Through Intelligence, Innovation, and Ethics. Semin Med Writ Educ [Internet]. 2025 Jan 1;4:35. Available from: https://mw.ageditor.ar/index.php/mw/article/view/35

8. Abdel Wahed M, Alqaraleh M, Salem Alzboon M, Subhi Al-Batah M. Application of Artificial Intelligence for Diagnosing Tumors in the Female Reproductive System: A Systematic Review. Multidiscip [Internet]. 2025 Jan 1;3:54. Available from: https://multidisciplinar.ageditor.uy/index.php/multidisciplinar/article/view/54

9. Rahman F, Chauhan R, Singh A, others. Texture-based abnormalities in oral squamous cell carcinoma detection using SVM. J Oral Cancer Res. 2018;10(3):85–92.

10. Kumar D, Gupta R, Mehta P. Automated segmentation of oral tissue layers using CNN and texture-based features. Int J Biomed Imaging. 2018;14(4):312–22.

11. Das S, Mitra S, Ghosh R. Quantitative assessment of keratinization in oral squamous cell carcinoma. J Cancer Diagnosis. 2015;7(2):45–52.

12. Patil R, Pawar S, Shinde V. Hybrid deep learning model for oral cancer detection using genetic algorithms. Int J Med Imaging. 2021;16(1):203–10.

13. Sarkar A, Chaudhary R, Mahajan S. Transfer learning-based framework for oral cancer histopathology classification. IEEE Trans Biomed Eng. 2019;66(5):1245–54.

14. Jain A, Verma R, Gupta S. Machine learning-based system for early detection of oral cancer using salivary biomarkers. J Med Diagnostics. 2020;12(1):52–60.

15. Alqaraleh M, Al-Batah M, Salem Alzboon M, Alzaghoul E. Automated Quantification of Vesicoureteral Reflux using Machine Learning with Advancing Diagnostic Precision. Data Metadata [Internet]. 2025 Jan 1;4:460. Available from: https://dm.ageditor.ar/index.php/dm/article/view/460

16. Al-Batah M, Salem Alzboon M, Alqaraleh M, Ahmad Alzaghoul F. Comparative Analysis of Advanced Data Mining Methods for Enhancing Medical Diagnosis and Prognosis. Data Metadata. 2024;3:465.

17. Abuashour A, Salem Alzboon M, Kamel Alqaraleh M, Abuashour A. Comparative Study of Classification Mechanisms of Machine Learning on Multiple Data Mining Tool Kits. Am J Biomed Sci Res 2024 [Internet]. 2024;22(1):577–9. Available from: www.biomedgrid.com

18. Alzboon MS, Al-Batah MS, Alqaraleh M, Abuashour A, Bader AFH. Early Diagnosis of Diabetes: A Comparison of Machine Learning Methods. Int J online Biomed Eng. 2023;19(15):144–65.

19. Al-Batah MS, Alzboon MS, Alzyoud M, Al-Shanableh N. Enhancing Image Cryptography Performance with Block Left Rotation Operations. Appl Comput Intell Soft Comput. 2024;2024(1):3641927.

20. Wahed MA, Alqaraleh M, Salem Alzboon M, Subhi Al-Batah M. Evaluating AI and Machine Learning Models in Breast Cancer Detection: A Review of Convolutional Neural Networks (CNN) and Global Research Trends. LatIA [Internet]. 2025 Jan 1;3:117. Available from: https://latia.ageditor.uy/index.php/latia/article/view/117

21. Alqaraleh M, Salem Alzboon M, Subhi Al-Batah M, Solayman Migdadi H. From Complexity to Clarity: Improving Microarray Classification with Correlation-Based Feature Selection. LatIA [Internet]. 2025 Jan 1;3:84. Available from: https://latia.ageditor.uy/index.php/latia/article/view/84

22. Al-Batah M, Zaqaibeh B, Alomari SA, Alzboon MS. Gene Microarray Cancer classification using correlation based feature selection algorithm and rules classifiers. Int J online Biomed Eng. 2019;15(8):62–73.

23. Alqaraleh M, Alzboon MS, Al-Batah MS, Wahed MA, Abuashour A, Alsmadi FH. Harnessing Machine Learning for Quantifying Vesicoureteral Reflux: A Promising Approach for Objective Assessment. Int J online Biomed Eng. 2024;20(11):123–45.

24. Al-Batah MS, Alzboon MS, Alazaidah R. Intelligent Heart Disease Prediction System with Applications in Jordanian Hospitals. Int J Adv Comput Sci Appl. 2023;14(9):508–17.

25. Alzboon MS. Internet of things between reality or a wishing - list : a survey. Int J Eng \& Technol. 2019;7(June):956–61.

26. Alzboon MS, Qawasmeh S, Alqaraleh M, Abuashour A, Bader AF, Al-Batah M. Machine Learning Classification Algorithms for Accurate Breast Cancer Diagnosis. In: 2023 3rd International Conference on Emerging Smart Technologies and Applications, eSmarTA 2023. 2023.

27. Alzboon MS, Aljarrah E, Alqaraleh M, Alomari SA. Nodexl Tool for Social Network Analysis. Turkish J Comput Math Educ. 2021;12(14):202–16.

28. Alqaraleh M, Salem Alzboon M, Mohammad SA-B. Optimizing Resource Discovery in Grid Computing: A Hierarchical and Weighted Approach with Behavioral Modeling. LatIA [Internet]. 2025 Jan 1;3:97. Available from: https://latia.ageditor.uy/index.php/latia/article/view/97

29. Alzboon MS, Al-Batah MS. Prostate Cancer Detection and Analysis using Advanced Machine Learning. Int J Adv Comput Sci Appl. 2023;14(8):388–96.

30. Alzboon MS, Qawasmeh S, Alqaraleh M, Abuashour A, Bader AF, Al-Batah M. Pushing the Envelope: Investigating the Potential and Limitations of ChatGPT and Artificial Intelligence in Advancing Computer Science Research. In: 2023 3rd International Conference on Emerging Smart Technologies and Applications, eSmarTA 2023. 2023.

31. Alqaraleh M, Salem Alzboon M, Subhi Al-Batah M. Real-Time UAV Recognition Through Advanced Machine Learning for Enhanced Military Surveillance. Gamification Augment Real [Internet]. 2025 Jan 1;3:63. Available from: https://gr.ageditor.ar/index.php/gr/article/view/63

32. Alzboon M. Semantic Text Analysis on Social Networks and Data Processing: Review and Future Directions. Inf Sci Lett. 2022;11(5):1371–84.

33. Alzboon MS. Survey on Patient Health Monitoring System Based on Internet of Things. Inf Sci Lett. 2022;11(4):1183–90.

34. Alzboon MS, Alomari S, Al-Batah MS, Alomari SA, Banikhalaf M. The characteristics of the green internet of things and big data in building safer, smarter, and sustainable cities Vehicle Detection and Tracking for Aerial Surveillance Videos View project Evaluation of Knowledge Quality in the E-Learning System View pr [Internet]. Vol. 6, Article in International Journal of Engineering and Technology. 2017. p. 83–92. Available from: https://www.researchgate.net/publication/333808921

35. Al Tal S, Al Salaimeh S, Ali Alomari S, Alqaraleh M. The modern hosting computing systems for small and medium businesses. Acad Entrep J. 2019;25(4):1–7.

36. Alzboon MS, Bader AF, Abuashour A, Alqaraleh MK, Zaqaibeh B, Al-Batah M. The Two Sides of AI in Cybersecurity: Opportunities and Challenges. In: Proceedings of 2023 2nd International Conference on Intelligent Computing and Next Generation Networks, ICNGN 2023. 2023.

37. Alomari SA, Alqaraleh M, Aljarrah E, Alzboon MS. Toward achieving self-resource discovery in distributed systems based on distributed quadtree. J Theor Appl Inf Technol. 2020;98(20):3088–99.

38. Alazaidah R. A Comparative Analysis of Discretization Techniques in Machine Learning. In: 2023 24th International Arab Conference on Information Technology, ACIT 2023. 2023. p. 1–6.

39. Alazaidah R, Owida HA, Alshdaifat N, Issa A, Abuowaida S, Yousef N. A comprehensive analysis of eye diseases and medical data classification. TELKOMNIKA (Telecommunication Comput Electron Control. 2024;22(6):1422–30.

40. Moubayed A, Injadat MN, Alhindawi N, Samara G, Abuasal S, Alazaidah R. A Deep Learning Approach Towards Student Performance Prediction in Online Courses: Challenges Based on a Global Perspective. In: 2023 24th International Arab Conference on Information Technology, ACIT 2023. 2023. p. 1–6.

41. Alazaidah R, Hassan M, Al-Rbabah L, Samara G, Yusof M, Al-Sherideh AS. Utilizing Machine Learning in Medical Diagnosis: Systematic Review and Empirical Analysis. In: 2023 24th International Arab Conference on Information Technology, ACIT 2023. 2023. p. 1–9.

42. Aziz DIABA, Yusoff M, Ibrahim N, Alazaidah R. Paddy Diseases Multi-Class Classification using CNN Variants. In: 2023 24th International Arab Conference on Information Technology, ACIT 2023. 2023. p. 1–8.

43. Alzyoud M, Alazaidah R, Aljaidi M, Samara G, Qasem MH, Khalid M, et al. Diagnosing diabetes mellitus using machine learning techniques. Int J Data Netw Sci. 2024;8(1):179–88.

44. Al-Batah MS, Al-Eiadeh MR. An improved binary crow-JAYA optimisation system with various evolution operators, such as mutation for finding the max clique in the dense graph. Int J Comput Sci Math. 2024;19(4):327-38.

45. Alzyoud M, Alazaidah R, Alzoubi H, Al-Shanableh N, Aljaidi M, Almatarneh S. Toward Identifying The Best Base Classifier in Multi Label Classification-an Investigative Study. In: 2023 24th International Arab Conference on Information Technology, ACIT 2023. 2023. p. 1–9.

46. Al-Batah MS. Modified recursive least squares algorithm to train the hybrid multilayered perceptron (HMLP) network. Appl Soft Comput. 2010;10(1):236-44.

47. Alazaidah R, Al-Qerem A, Qasem MH, Al-Shaikh A, Almilli N, Injadat MN. Feature Selection in Associative Classification-A Review and Comparative Analysis. In: 2023 24th International Arab Conference on Information Technology, ACIT 2023. 2023. p. 1–5.

38. Al-Batah MS. Testing the probability of heart disease using classification and regression tree model. Annu Res Rev Biol. 2014;4(11):1713-25.

49. Qasem MH, Aljaidi M, Samara G, Alsarhan A, Alazaidah R, Ali Al-Gumaei YO, et al. Towards Advancing Distributed Data Mining: Intelligent Agent Systems. In: 2nd International Engineering Conference on Electrical, Energy, and Artificial Intelligence, EICEEAI 2023. 2023. p. 1–5.

50. Al-Batah MS. Integrating the principal component analysis with partial decision tree in microarray gene data. IJCSNS Int J Comput Sci Netw Secur. 2019;19(3):24-29.

51. Al-Batah MS, Al-Eiadeh MR. An improved discreet Jaya optimisation algorithm with mutation operator and opposition-based learning to solve the 0-1 knapsack problem. Int J Math Oper Res. 2023;26(2):143-69.

52. Alazaidah R, Samara G, Aljaidi M, Haj Qasem M, Alsarhan A, Alshammari M. Potential of Machine Learning for Predicting Sleep Disorders: A Comprehensive Analysis of Regression and Classification Models. Diagnostics. 2024;14(1):27.

53. Al-Batah MS. Ranked features selection with MSBRG algorithm and rules classifiers for cervical cancer. Int J Online Biomed Eng. 2019;15(12):4.

**FINANCING**

This work is supported from Jadara University under grant number [Jadara-SR-Full2023], and Zarqa University.

**CONFLICT OF INTEREST**

The authors declare that the research was conducted without any commercial or financial relationships that could be construed as a potential conflict of interest.

**AUTHORSHIP CONTRIBUTION**

*Conceptualization:* Muhyeeddin Alqaraleh, Mowafaq Salem Alzboon.
*Data Curation:* Mohammad Al-Batah.
*Formal Analysis:* Muhyeeddin Alqaraleh, Mohammad Al-Batah.
*Research:* Muhyeeddin Alqaraleh, Mowafaq Salem Alzboon, Mohammad Al-Batah.
*Methodology:* Muhyeeddin Alqaraleh, Mohammad Al-Batah.
*Project Management:* Mowafaq Salem Alzboon, Mohammad Al-Batah.
*Resources:* Mowafaq Salem Alzboon, Mohammad Al-Batah.
*Software:* Mowafaq Salem Alzboon, Mohammad Al-Batah.
*Supervision:* Mowafaq Salem Alzboon, Mohammad Al-Batah.
*Validation:* Muhyeeddin Alqaraleh, Mohammad Al-Batah.
*Display:* Mowafaq Salem Alzboon, Mohammad Al-Batah.
*Drafting - Original Draft:* Muhyeeddin Alqaraleh, Mohammad Al-Batah.
*Writing - Proofreading and Editing:* Muhyeeddin Alqaraleh, Mowafaq Salem Alzboon, Mohammad Al-Batah.